\documentclass[letterpaper, 10 pt, conference]{ieeeconf}
\IEEEoverridecommandlockouts
\usepackage{amsmath,amssymb,amsfonts}
\usepackage{graphicx}
\usepackage{caption}
\usepackage{subcaption}
\usepackage{cite}

\begin{document}

\title{Physically Consistent Online Inertial Adaptation for \\ Humanoid Loco-manipulation
\thanks{$^{1}$Intelligent Systems \& Robotics, University of West Florida, USA}
\thanks{$^{2}$Robotics Lab, Florida Institute for Human \& Machine Cognition, USA}
\thanks{$^{3}$Boardwalk Robotics, USA}
\thanks{\newline \texttt{jf127@students.uwf.edu, \newline \{smccrory, sbertrand, rgriffin\}@ihmc.org, \newline christian.debuys@boardwalkrobotics.com}}
}

\author{James Foster$^{1}$, Stephen McCrory$^{1,2}$, Christian DeBuys${^3}$, Sylvain Bertrand${^2}$, and Robert Griffin$^{1,2,3}$}

\maketitle

\begin{abstract}
The ability to accomplish manipulation and locomotion tasks in the presence of significant time-varying external loads is a remarkable skill of humans that has yet to be replicated convincingly by humanoid robots. Such an ability will be a key requirement in the environments we envision deploying our robots: dull, dirty, and dangerous. External loads constitute a large model bias, which is typically unaccounted for. In this work, we enable our humanoid robot to engage in loco-manipulation tasks in the presence of significant model bias due to external loads. We propose an online estimation and control framework involving the combination of a physically consistent extended Kalman filter for inertial parameter estimation coupled to a whole-body controller. We showcase our results both in simulation and in hardware, where weights are mounted on Nadia's wrist links as a proxy for engaging in tasks where large external loads are applied to the robot.
\end{abstract}


\section{Introduction}

State-of-the-art model-based controllers rely heavily on an internal dynamic model, usually used inside an optimization routine to repeatedly plan the best sequence of control actions given the current (estimated) state of the robot. However, most current model-based approaches in humanoid locomotion and manipulation utilize a dynamic model that is identified offline but remains unchanged online. This introduces several difficulties. Firstly, there is often significant engineering time spent tuning the dynamic model to compensate for inevitable model mismatch. While one would hope that a good controller is robust to some degree of model mismatch, a second and more practical issue is that a frozen dynamic model restricts the robot to performing manipulation and locomotion tasks where the model is known to some acceptable margin. This often results in manipulation tasks involving objects of trivial mass relative to the robot, and precludes locomotion tasks where any significant load is placed on the robot.

\begin{figure}
    \centering
    \includegraphics[width=7.0cm,height=!]{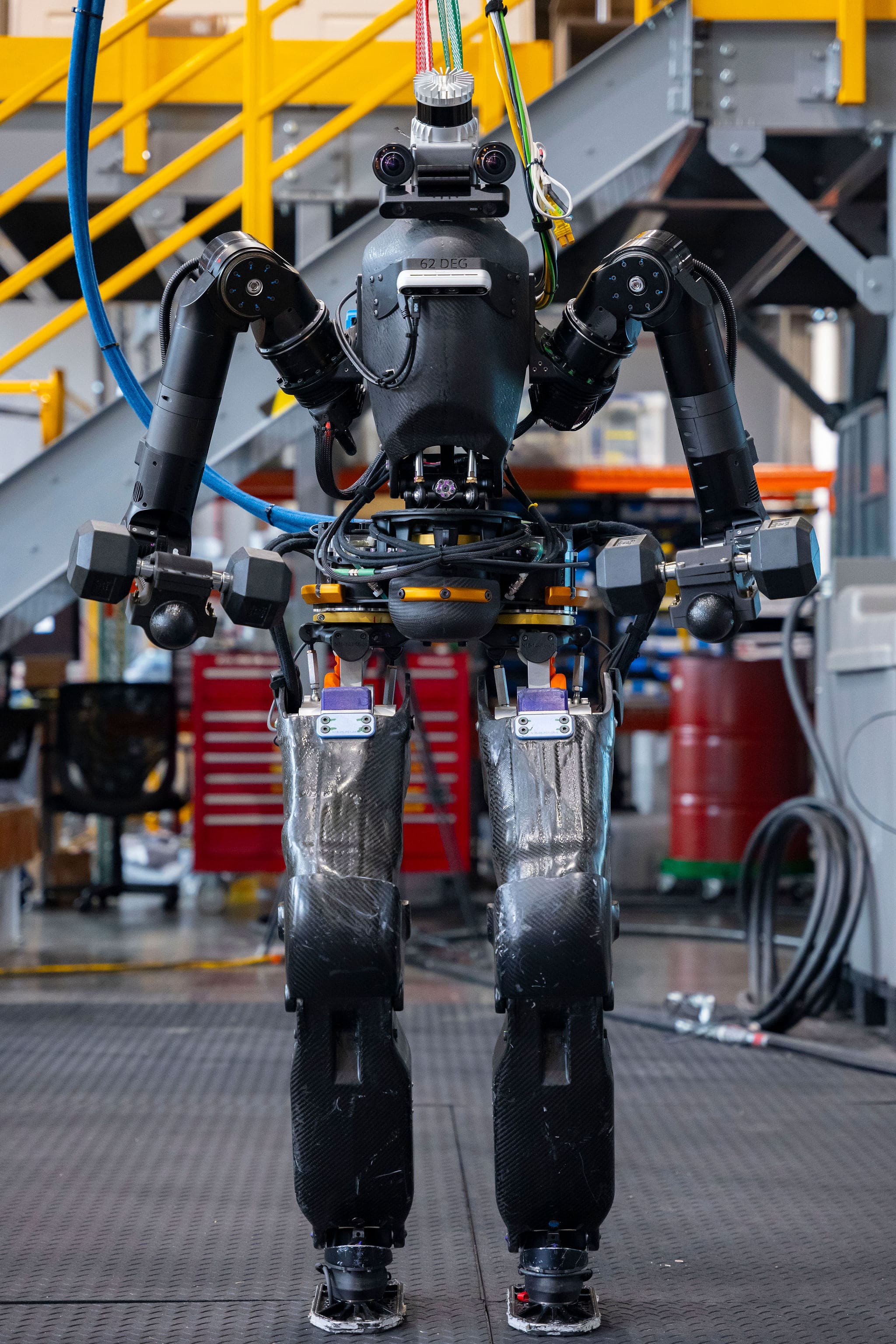}
    \caption{Our humanoid robot Nadia actively estimating and adapting to 12lb dumbbells attached to the forearms. We use these weights as a proxy for inertially significant manipulation tasks, such as heavy tool use.}
    \label{fig:money}
\end{figure}

These restrictions stand in stark contrast to humans and other animals, which have been experimentally shown to adapt their motor behaviour in the presence of consistent model bias \cite{orban2011representations, mcnamee2019internal}. For humanoids to engage in meaningful, non-trivial interaction with the environment, they must be capable of identifying changes in their internal dynamic model and updating it for use in downstream control routines. In this paper, we enable identification and adaptation in the presence of large external loads on our humanoid robot Nadia, as seen in Fig.~\ref{fig:money}, by proposing a framework consisting of an extended Kalman filter (EKF) that streams inertial parameter estimates to a whole-body controller online. The proposed filter is lightweight and general in that it can be easily modified to estimate the parameters of any link in a given rigid body system, removing the dependence on a reference model and thus making it controller-agnostic. Furthermore, the proposed filter builds upon a recently introduced ``physically consistent'' inertial parameterization~\cite{rucker2022smooth}, guaranteeing that the streamed estimates are physically realizable and precluding catastrophic errors in the controller.

\subsection{Related Work}

There are several excellent recent surveys on inertial identification and model learning. In \cite{leboutet2021inertial}, various concrete least-squares formulations, Kalman filtering approaches, neural network methods, and semidefinite programming approaches are compared, whereas \cite{geist2021structured} provides a more conceptual overview of the exploitable structure of rigid body dynamics and how to couple this structure with data-driven modeling techniques. Finally, \cite{lee2023robot} covers different metrics, sources of error, and recent advances in geometric methods for parameter identification.

Most inertial parameter estimation works leverage the finding that the inverse dynamics are linear with respect to inertial parameters \cite{atkeson1986estimation}. Subsequent works have highlighted the additional structure the inertial parameters must obey \cite{yoshida1996set, traversaro2016identification, wensing2017linear}: beyond the obvious constraint on positive mass, the moment of inertia matrix must satisfy additional geometric conditions like positive definiteness and the principal moments of inertia must satisfy a triangle inequality. Unfortunately, for multi-link systems like manipulators, particularly for high degree of freedom systems like humanoids, uniquely identifying inertial parameters becomes challenging \cite{ayusawa2014identifiability}. A lack of identifiability can be due to the kinematic structure of the robot \cite{bellman1970structural} or from a lack of persistent excitation in the data \cite{gautier1992exciting}, but methods exist to characterize the identifiable parameters either through numerical \cite{gautier1989identification} or analytical \cite{wensing2017observability} means. Extensive work has gone into methods for generating rich and informative trajectories for robot parameter identification \cite{swevers1997optimal} with several applications to humanoids \cite{lee2022optimized, jovic2016humanoid, bonnet2018inertial}. Alternatively, several control-oriented works on humanoids \cite{mori2018online, rotella2015humanoid} have estimated gross robot quantities like center of mass and momentum online.

There are also adaptive control approaches that update parameters based on task error. The seminal work \cite{slotine1987adaptive} has been extended in many ways: from the fully actuated to the underactuated case \cite{pucci2015collocated}, modifying the adaptation law to account for the geometric properties of the inertial parameters \cite{lee2018natural}, and embedding the adaptation law as a control-Lyapunov function inequality constraint in a model predictive controller \cite{minniti2021adaptive}. Recent work in \cite{sombolestan2023adaptive} applies L1 adaptive control to quadrupedal load carrying, another direct adaptive control approach with an adaptation law based on the difference in behavior between a reference model and the actual robot. In \cite{elobaid2023online}, a centroidal nonlinear MPC scheme on iCub is proposed where the attenuation of an external wrench from a payload is included as a task.

\subsection{Contributions}

From our survey of existing work, we note that most approaches to parameter estimation take place offline in a system identification context, with notable online exceptions for estimating gross robot quantities like center of mass or momentum, which is useful for the locomotion task but less so for other tasks. Online task-error approaches require tight coupling with an existing reference model or controller, reducing their flexibility and complicating the overall design process. In this paper, we contribute:
\begin{itemize}
    \item A lightweight, controller-agnostic extended Kalman filtering framework that guarantees physically consistent parameter estimates online. 
    \item A compact problem formulation that assumes only a subset of the inertial parameters will be estimated online, resulting in more efficient rigid body dynamics calculations inside the filter.
    \item Verification of the proposed filter in simulation, showing the benefit of physical consistency over a baseline Kalman filter, and improvements in control performance when our filter is included in an adaptive control framework versus a controller with no adaptation.
    \item Demonstrations on hardware of the robot adapting to loads on the forearm links, and an innovation gating method for rejecting undesirable transients in parameter estimates due to contact impulses during locomotion.
\end{itemize}
\section{Background}

We assume that the robot can be adequately modelled as a floating base rigid body system, with the equations of motion given by:
\begin{equation}
\mathbf{M}(\mathbf{q})\mathbf{\dot{v}} + \mathbf{c}(\mathbf{q}, \mathbf{v}) = \mathbf{S}^\top \boldsymbol{\tau} + \sum_{i \in \mathcal{C}}\mathbf{J}_i^\top(\mathbf{q}) \boldsymbol{\lambda}_i,
\end{equation}
where $\mathbf{q} \in \mathbb{R}^{n_q}$, $\mathbf{v} \in \mathbb{R}^{n_v}$ are the generalized configuration and velocity of the robot, $\boldsymbol{\tau} \in \mathbb{R}^{n_u}$ are the joint torques of the robot, and $\mathbf{S}$ is the selection matrix for the actuated degrees of freedom. The pair $\mathbf{J}_i$ and $\boldsymbol{\lambda}_i$ are the Jacobian and wrench corresponding to the $i$th contact in the set of external contacts $\mathcal{C}$. $\mathbf{M}$ is the mass matrix and $\mathbf{c}$ collects the centripetal, Coriolis, and gravitational terms.

Denoting the number of rigid bodies in the system by $N$, we can follow \cite{mistry2009inertial} and write the inverse dynamics in a form where the inertial parameters of the robot appear in a linear fashion:
\begin{equation} \label{eq:invdyn-regressor}
\mathbf{Y}(\mathbf{q}, \mathbf{v}, \mathbf{\dot{v}})\boldsymbol{\pi} = \mathbf{S}^\top \boldsymbol{\tau} + \sum_{i \in \mathcal{C}}\mathbf{J}_i^\top(\mathbf{q}) \boldsymbol{\lambda}_i,
\end{equation}
where $\mathbf{Y} \in \mathbb{R}^{n_v \times 10N}$ is a state-dependent regressor matrix that maps the stacked inertial parameters $\boldsymbol{\pi} = [\boldsymbol{\pi}_1^\top, \boldsymbol{\pi}_2^\top, \dots, \boldsymbol{\pi}_N^\top]^\top \in \mathbb{R}^{10N}$ into torque. The inertial parameters for the $n^\text{th}$ individual body $\boldsymbol{\pi}_n \in \mathbb{R}^{10}$ consist of its mass $m$, first moment of mass $\mathbf{h} \in \mathbb{R}^3$, and the unique elements of the symmetric rotational inertia matrix $\mathbf{I} \in \mathbb{R}^{3 \times 3}$:
\begin{equation}
\boldsymbol{\pi}_n = [m, h_{x}, h_{y}, h_{z}, I_{xx}, I_{xy}, I_{xz}, I_{yy}, I_{yz}, I_{zz}]^\top.
\end{equation}
As in \cite{mistry2009inertial}, it is then possible to collect samples of robot motion and leverage the linear structure of the equations of motion to get parameter estimates by solving a least squares problem. Concretely, one gathers $K$ samples of $\mathbf{q}$, $\mathbf{v}$, $\mathbf{\dot{v}}$, $\boldsymbol{\tau}$, and $\boldsymbol{\lambda}$, and creates the following stacked problem data:
\begin{equation} \label{eq:stacked}
\underbrace{\begin{bmatrix}
\mathbf{Y}_1 \\
\mathbf{Y}_2 \\
\vdots \\
\mathbf{Y}_K
\end{bmatrix}}_{\mathbf{\bar{Y}}} \boldsymbol{\pi} = 
\underbrace{\begin{bmatrix}
\mathbf{S}^\top \boldsymbol{\tau}_1 + \sum_i \mathbf{J}_1^\top \boldsymbol{\lambda}_1 \\
\mathbf{S}^\top \boldsymbol{\tau}_2 + \sum_i \mathbf{J}_2^\top \boldsymbol{\lambda}_2 \\
\vdots \\ 
\mathbf{S}^\top \boldsymbol{\tau}_K + \sum_i \mathbf{J}_K^\top \boldsymbol{\lambda}_K
\end{bmatrix}}_{\mathbf{\bar{f}}}
\end{equation}
where $\sum_i \mathbf{J}_j^\top \boldsymbol{\lambda}_j$ is an abuse of notation denoting the sum of contact wrench torque contributions over contacts $i \in \mathcal{C}$ for sample $j$. One then solves the linear least squares problem:
\begin{equation} \label{eq:least-squares}
\min_{\boldsymbol{\hat{\pi}}} \tfrac{1}{2} ||\mathbf{\bar{f}} - \mathbf{\bar{Y}} \boldsymbol{\hat{\pi}} ||_\mathbf{W}^2,
\end{equation}
where $\mathbf{W}$ is an appropriately chosen weight matrix that captures the variance of the measurements and the characteristic scales of the different units (e.g. mass vs. inertia).

Though we do not follow the batch estimation procedure in (\ref{eq:least-squares}), we include it to highlight several difficulties of the problem (see also \cite{leboutet2021inertial}):

\textit{Unobservability due to kinematic structure.} Oftentimes the kinematic structure of the robot renders certain parameters unobservable, or observable only in linear combinations \cite{bellman1970structural}. This will cause $\mathbf{\bar{Y}}$ to become rank-deficient.

\textit{Lack of persistent excitation in motion data.} If the motion data is not sufficiently rich enough to excite all the degrees of freedom, again $\mathbf{\bar{Y}}$ will be rank-deficient. In both cases, (\ref{eq:least-squares}) can be regularized and solved by methods such as damped least squares \cite{atkeson1986estimation} or other regularization techniques \cite{lee2019geometric}.

\textit{Physically inconsistent estimates.} While (\ref{eq:least-squares}) exploits the linearity of the inverse dynamics with respect to inertial parameters, it does not take into account additional problem structure \cite{yoshida1996set, traversaro2016identification}; depending on the data, one may receive estimates where mass is negative, and moment of inertia matrices that are physically impossible. If a parameter estimation algorithm is run in a live control loop, as in this work, non-physical parameter estimates could yield catastrophic results when used in a downstream model-based controller.
\section{Method}

\subsection{Condensing the inertial estimation problem}

Most extant works on inertial parameter estimation for humanoids have focused on system identification, where most (if not all) of the inertial parameters are considered to be poorly known and thus required to estimate.

In this work, we are more focused on online estimation and adaptation. We make the assumption that a prior offline system identification procedure has been performed, so that many of the inertial parameters are well-known and can be considered fixed. Furthermore, we assume knowledge of the tasks the robot will perform, such that we know what links will vary inertially. Therefore, only a subset of the robot's inertial parameter vector $\boldsymbol{\pi}$ needs to be actively estimated online. Indeed, one can split the product of the regressor matrix and parameter vector into ``estimate'' $\boldsymbol{*}^\text{est}$ and ``nominal'' $\boldsymbol{*}^\text{nom}$ terms:
\begin{equation} \label{eq:decomp}
\mathbf{Y}\boldsymbol{\pi} = \mathbf{Y}^\text{est}\boldsymbol{\pi}^\text{est} + \mathbf{Y}^\text{nom} \boldsymbol{\pi}^\text{nom},
\end{equation}
where typically $\text{dim}(\boldsymbol{\pi}^\text{est}) << \text{dim}(\boldsymbol{\pi}^\text{nom})$.

This decomposition has multiple benefits. Firstly, it can help to resolve linear dependence issues by pinning the inertial parameters in $\boldsymbol{\pi}^\text{nom}$ to certain values. Secondly, one can exploit some well-known concepts in the rigid body dynamics literature to efficiently calculate the two terms in (\ref{eq:decomp}). For the first term, $\mathbf{Y}^\text{est}$ can be found via a modified recursive Newton-Euler algorithm (RNEA), where the forward pass is run just once independently, and spatial forces in the backward pass are linearly parameterized with respect to the inertial parameters, such an approach is implemented in multiple popular software libraries \cite{pinocchioweb, spatialv2extended}. For the second term, one can use the assumed-known $\boldsymbol{\pi}^\text{nom}$: the equivalent torque $\boldsymbol{\tau}^\text{nom}$ can be computed with a \textsl{single} RNEA call where any inertial parameters not in $\boldsymbol{\pi}^\text{nom}$ are zeroed. This concept is similar to that mentioned in \cite[p. 94]{featherstone2014rigid}, where the zeroing of certain parameters in RNEA yields certain dynamic quantities. These simplifications result in:
\begin{equation} \label{eq:decomp-tau}
    \mathbf{Y}\boldsymbol{\pi} = \mathbf{Y}^\text{est}\boldsymbol{\pi}^\text{est} + \boldsymbol{\tau}^\text{nom}.
\end{equation}

We next show this efficient regressor calculation can be used as the measurement model in a Kalman Filter designed to estimate $\boldsymbol{\pi}^\text{est}$.

\subsection{Inertial Kalman Filter} \label{sec:kf}

Following \cite[Chapter 5.1]{simon2006optimal}, the standard form for the process model and measurement model of a discrete-time Kalman filter with no control input is:
\begin{align}
    \label{eq:process-standard}
    \mathbf{x}_k &= \mathbf{F}_k \mathbf{x}_{k-1} + \boldsymbol{\eta}_k, \quad &\boldsymbol{\eta}_k \sim \mathcal{N}(\mathbf{0}, \mathbf{Q}_k), \\
    \label{eq:measurement-standard}
    \mathbf{z}_k &= \mathbf{H}_k \mathbf{x}_k + \boldsymbol{\nu}_k, \quad &\boldsymbol{\nu}_k \sim \mathcal{N}(\mathbf{0}, \mathbf{R}_k).
\end{align}
Within a single time instant $k$, the \textsl{predict} and \textsl{update} step proceed as:
\begin{align}
\begin{split} \label{eq:predict}
    \text{predict:} \quad
    \mathbf{x}_k^- &= \mathbf{F}_k \mathbf{x}_{k-1} \\
    \mathbf{P}_k^- &=\mathbf{F}_k \mathbf{P}_{k-1} \mathbf{F}_k^\top + \mathbf{Q}_k \\
\end{split}
\end{align}
\begin{align}
\begin{split} \label{eq:update}
    \text{update:} \quad
    \tilde{\mathbf{y}}_k &= \mathbf{z}_k - \mathbf{H}_k \mathbf{x}_k^- \\
    \mathbf{S}_k &= \mathbf{H}_k \mathbf{P}_k^- \mathbf{H}_k^\top + \mathbf{R}_k \\
    \mathbf{K}_k &= \mathbf{P}_k^- \mathbf{H}_k^\top \mathbf{S}_k^{-1} \\
    \mathbf{x}_k^+ &= \mathbf{x}_k^- + \mathbf{K}_k \tilde{\mathbf{y}}_k \\
    \mathbf{P}_k^+ &= (\mathbf{I} - \mathbf{K}_k \mathbf{H}_k) \mathbf{P}_k^-
\end{split}
\end{align}

Concerning the process model for inertial parameter estimation, we follow existing work in the literature (see \cite{rotella2015humanoid} for a humanoid example, and \cite[Chapter 5.1]{barfoot2024state} for a more general discussion of estimating constant or slowly varying parameters) by assuming the only change in the parameters being estimated is driven by the white noise $\boldsymbol{\eta}_k$:
\begin{equation} \label{eq:process}
    \boldsymbol{\pi}_k^\text{est} = \boldsymbol{\pi}_{k-1}^\text{est} + \boldsymbol{\eta}_k.
\end{equation}

To form the measurement, we plug (\ref{eq:decomp-tau}) into (\ref{eq:invdyn-regressor}), replacing $\mathbf{S}^\top \boldsymbol{\tau}$ with the observed torque from the robot $\boldsymbol{\tau}^\text{obs}$:
\begin{equation}
    \mathbf{Y}^\text{est}\boldsymbol{\pi}^\text{est} + \boldsymbol{\tau}^\text{nom} =
        \boldsymbol{\tau}^\text{obs} + \sum_{i \in \mathcal{C}} \mathbf{J}^\top_i \boldsymbol{\lambda}_i
\end{equation}
We then introduce the variable $\Delta \boldsymbol{\tau}$ to isolate the contribution from the estimated parameters: 
\begin{equation} \label{eq:delta-tau}
    \Delta \boldsymbol{\tau} = \boldsymbol{\tau}^\text{obs} - \boldsymbol{\tau}^\text{nom} + \sum_{i \in \mathcal{C}} \mathbf{J}^\top_i \boldsymbol{\lambda}_i,
\end{equation}
which leaves $\Delta \boldsymbol{\tau} \approx \mathbf{Y}^\text{est} \boldsymbol{\pi}^\text{est}$, justifying the use of the measurement model:
\begin{equation} \label{eq:measurement}
    \Delta \boldsymbol{\tau}_k = \mathbf{Y}_k^\text{est} \boldsymbol{\pi}^\text{est}_k + \boldsymbol{\nu}_k.
\end{equation}

\begin{figure*}[ht]
\vspace{0.5\baselineskip}
    \centering
\includegraphics[width=0.98\linewidth,height=!]{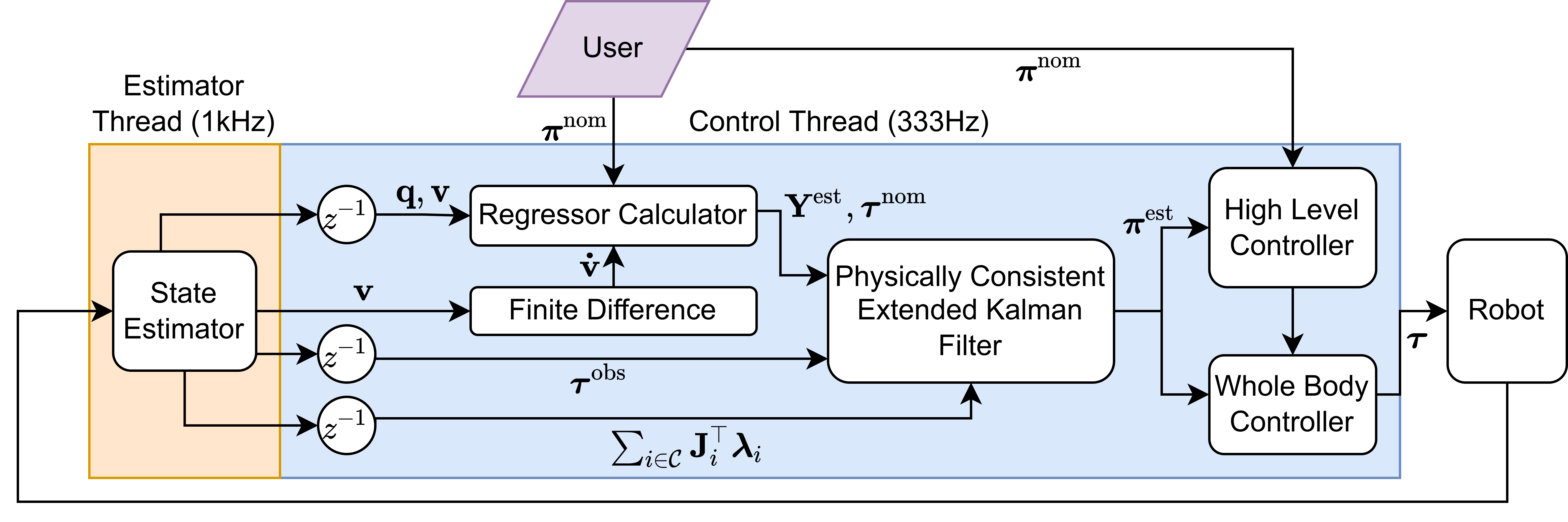}
    \caption{Block diagram of proposed filter and its interface with the controller, as described in Section \ref{sec:integration}.}
    \label{fig:block-diagram}
\end{figure*}

\subsection{Physically Consistent Inertial Extended Kalman Filter}

The linear least squares problem in (\ref{eq:least-squares}) only considers the inertial parameters as members of the Euclidean space $\mathbb{R}^{10}$. However, there are additional conditions that the inertial parameters must satisfy. \cite{yoshida1996set} introduces the conditions of the mass $m$ being positive and the rotational inertia matrix being positive definite. These conditions have since been termed \textsl{physical consistency}. Subsequently, \cite{traversaro2016identification} noted that the rotational inertia matrix being positive definite was a necessary but not sufficient condition. Coining the term \textsl{full physical consistency}, \cite{traversaro2016identification} shows in addition that a set of triangle inequalities between the principal moments of inertia must also be satisfied. Their solution method involves manifold optimization constraints, which are difficult to utilize in an online real-time control framework. Recently, \cite{rucker2022smooth} introduced an inertial parameterization that is fully physically consistent by construction, revolving around a log-Cholesky decomposition of the pseudo-inertia matrix of a rigid body. This new parameterization for the $n^\text{th}$ rigid body is given by $\boldsymbol{\theta}_n \in \mathbb{R}^{10}$: 
\begin{equation}
\boldsymbol{\theta}_n = [\alpha, d_1, d_2, d_3, s_{12}, s_{13}, s_{23}, t_1, t_2, t_3]^\top,
\end{equation}
where $\alpha$ scales the density of a reference body, and $d$, $s$, and $t$ respectively provide a scaling, shearing, and translation of a reference body. One can map from the log-Cholesky parameterization to the default parameterization:
\begin{equation} \label{eq:log-cholesky}
\boldsymbol{\pi}_n = \mathbf{g}(\boldsymbol{\theta}_n) =
e^{2 \alpha}
\begin{bmatrix}
t_1^2 + t_2^2 + t_3^2 + 1 \\
t_1 e^{d_1} \\
t_1 s_{12} + t_2 e^{d_2} \\
t_1 s_{13} + t_2 s_{23} + t_3 e^{d_3} \\
s_{12}^2 + s_{13}^2 + s_{23}^2 + e^{2d_2} + e^{2d_3} \\
-s_{12} e^{d_1} \\
-s_{13} e^{d_1} \\
s_{13}^2 + s_{23}^2 + e^{2d_1} + e^{2d_3} \\
-s_{12} s_{13} - s_{23} e^{d_2} \\
s_{12}^2 + e^{2 d_1} + e^{2 d_2}
\end{bmatrix}
\end{equation}
where the ordering is slightly different from \cite{rucker2022smooth} due to the use of a different ordering of $\boldsymbol{\pi}$ in this work.

As shown in \cite{rucker2022smooth}, the mapping $\mathbf{g}$ is smooth, unique, and invertible, allowing one to easily move back and forth between parameterizations as required. Furthermore, the Jacobian of $\mathbf{g}$ is closed-form, which we can exploit in our extended Kalman filtering framework. We define a similar process model to (\ref{eq:process}), where the parameters to estimate are assumed constant, but in the log-Cholesky parameterization:
\begin{equation} \label{eq:process-log-cholesky}
    \boldsymbol{\theta}^\text{est}_k = \boldsymbol{\theta}^\text{est}_{k-1} + \boldsymbol{\eta}_k,
\end{equation}
with the corresponding measurement model:
\begin{equation} \label{eq:measurement-log-cholesky}
    \Delta \boldsymbol{\tau}_k = \underbrace{\mathbf{Y}^\text{est}_k\mathbf{g}(\boldsymbol{\theta}^\text{est}_k)}_{\mathbf{h}(\boldsymbol{\theta}^\text{est}_k)} + \boldsymbol{\nu}_k.
\end{equation}
Introducing the nonlinear measurement model $\mathbf{h}(\boldsymbol{\theta}^\text{est})$ requires the modification of the first two computations of the standard Kalman filter update step in (\ref{eq:update}). For the inertial estimation problem at hand, we have:
\begin{align}
    \begin{split}
      \Delta \tilde{\boldsymbol{\tau}}_k &= \Delta \boldsymbol{\tau}_k - \mathbf{h}(\boldsymbol{\theta}^{\text{est}-}_k), \\
      \mathbf{S}_k &= \underbrace{(\mathbf{Y}^\text{est}_k \mathbf{G}_k)}_{\mathbf{H}_k} \mathbf{P}_k^- \underbrace{(\mathbf{Y}^\text{est}_k \mathbf{G}_k)^\top}_{\mathbf{H}_k^\top} + \mathbf{R}_k,
    \end{split}
\end{align}
\begin{equation}
    \mathbf{G}_k = \frac{\partial \mathbf{g}}{\partial \boldsymbol{\theta}}\Big|_{\boldsymbol{\theta}_k^{\text{est}-}}
    \ \ \text{therefore} \ \
    \mathbf{H}_k = \frac{\partial \mathbf{h}}{\partial \boldsymbol{\theta}}\Big|_{\boldsymbol{\theta}_k^{\text{est}-}} = \mathbf{Y}^\text{est}_k \mathbf{G}_k 
\end{equation}
To summarise, these modifications require the forfeit of the linear least squares structure, mandating a move from the regular Kalman filter to the extended Kalman filter, but result in inertial parameters that are fully physically consistent by construction at every stage of the solve procedure.

\subsection{Integration with state estimator and controller} \label{sec:integration}

 A block diagram showing the structure of our approach is given in Fig. \ref{fig:block-diagram}. We run the inertial parameter estimator inside the control thread that is pinned at 333Hz. We read configuration $\mathbf{q}$, velocity $\mathbf{v}$, observed joint torques $\boldsymbol{\tau}^\text{obs}$, and contact wrench information $\boldsymbol{\lambda}$ from the state estimator that runs at 1kHz. Accelerations $\mathbf{\dot{v}}$ are computed inside the estimator via finite differences, and the other signals are delayed to compensate for the lag introduced. The user supplies nominal parameters $\boldsymbol{\pi}^\text{nom}$, and thus implicitly describes the parameters to be estimated $\boldsymbol{\pi}^\text{est}$. The physically consistent EKF internally uses (\ref{eq:process-log-cholesky}) and (\ref{eq:measurement-log-cholesky}) before converting the estimate back to the standard inertial parameterization $\boldsymbol{\pi}^\text{est}$ which is passed through to the high-level controller and whole-body controller. The high-level controller is responsible for specifying motion tasks that are passed as objectives to the quadratic program solved in the whole-body controller \cite{koolen2016design}.
 
\section{Results}

\subsection{Simulation}

\begin{figure}[t!]
    \centering
    \includegraphics[width=0.98\linewidth,height=!]{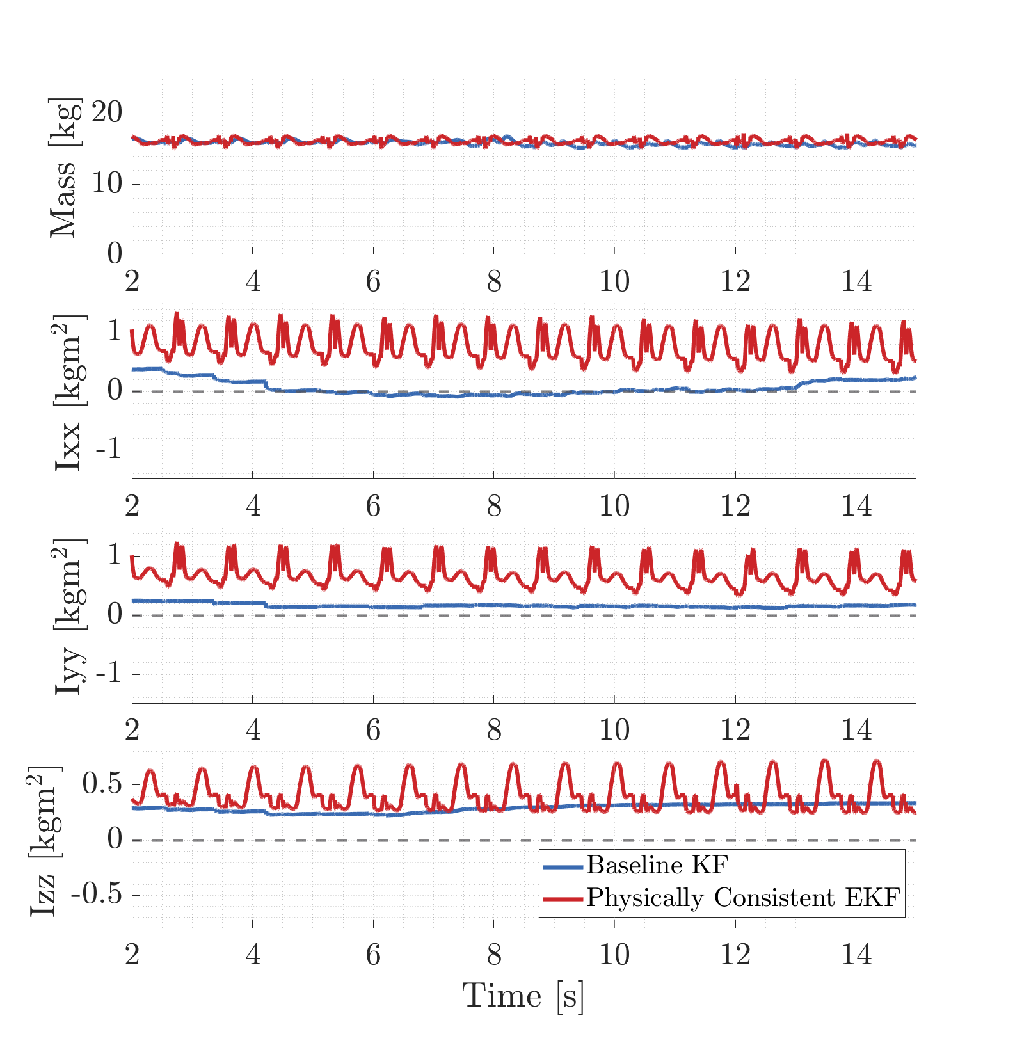}
    \caption{Parameter estimates for the robot's torso link from the baseline Kalman filter and the Log-Cholesky extended Kalman filter.}
    \label{fig:physically-consistent}
\end{figure}

We first showcase the benefit of the physically consistent Log-Cholesky parameterization from \cite{rucker2022smooth} in the Extended Kalman Filter (\ref{eq:process-log-cholesky}) and (\ref{eq:measurement-log-cholesky}) over a baseline Kalman Filter implementation (\ref{eq:process}) and (\ref{eq:measurement}). Fig. \ref{fig:physically-consistent} shows the estimates of our robot's torso link over time when the robot is instructed to walk forward. Whilst both estimators recover the torso mass, the lack of excitation for identifying the inertia diagonals causes the baseline Kalman filter estimates for $I_{xx}$, $I_{yy}$, and $I_{zz}$ to drift and become physically inconsistent almost immediately, with $I_{xx}$ becoming negative for a short time interval. In contrast, the inertia diagonals for the physically consistent EKF vary periodically, but always remain physically consistent.

\begin{table}[b!]
\centering
\caption{Simulation mass variation waveforms.}
\label{tab:waves}
\renewcommand{\arraystretch}{1.1}
\begin{tabular}{||c|c|c|c|c||}
\hline
Link & Waveform & Offset & Amplitude & Frequency \\
\hline
\hline
Torso & Square & 30kg & 5kg & 0.2Hz \\ 
Forearm & Sine/Square & 4.5kg & 3kg & 1Hz  \\ 
\hline
\end{tabular}
\end{table}

\begin{figure}[t!]
    \centering
    \vspace{2\baselineskip}
\includegraphics[width=0.9\linewidth,height=!]{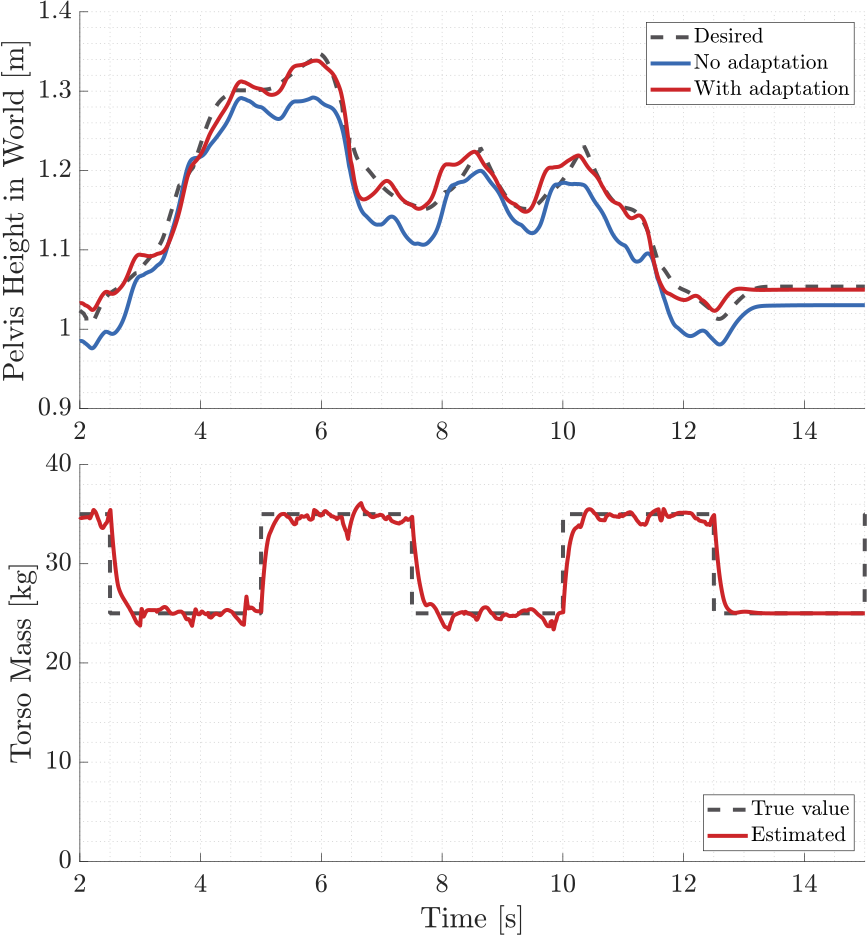}
    \caption{Pelvis height task performance over rough terrain, with and without adaptation of the torso inertial parameters.}
    \label{fig:pelvis-adaptation}
\end{figure}

\begin{figure}[t!]
    \centering
\includegraphics[width=0.90\linewidth,height=!]{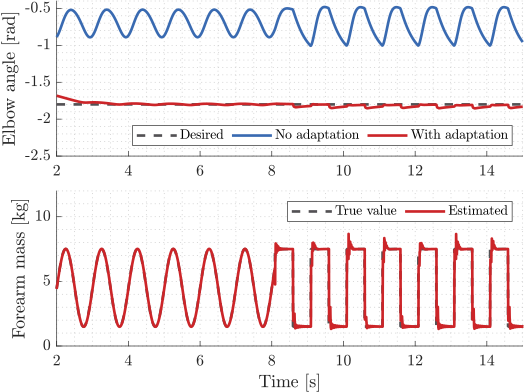}
    \caption{Elbow jointspace task performance with and without adaptation of the forearm link inertial parameters. Around the 8s mark, the mass waveform is changed from a sine wave to the square wave with the same properties.}
    \label{fig:elbow-adaptation}
\end{figure}

We next show the benefit of integrating our inertia estimation pipeline with a downstream whole-body controller in an adaptive control manner. In Fig. \ref{fig:pelvis-adaptation}, we compare the performance on a pelvis height control task with and without adaptation of the torso link's inertial parameters, where the mass of the torso link is varied to simulate a load-carrying task. In Fig. \ref{fig:elbow-adaptation}, we compare the performance on an elbow jointspace control task with and without adaptation of the forearm link's inertial parameters. The mass of the forearm link is varied to simulate the manipulation of a heavy tool. The waveforms used for both tasks are detailed in Table \ref{tab:waves}.


\begin{table}[h!]
\vspace{0.5\baselineskip}
\centering
\caption{Root-mean-squared error for simulated tasks.}
\label{tab:rmse}
\renewcommand{\arraystretch}{1.1}
\begin{tabular}{||c|c|c||}
\hline
Task & No adaptation & With adaptation \\
\hline
\hline
Pelvis height & 0.0352m & 0.0157m \\ 
Elbow angle & 1.0956rad & 0.1499rad \\ 
\hline
\end{tabular}
\end{table}

For the pelvis height control task, it can be observed that the controller tracks the desired pelvis height much more closely with an adaptive estimate of the torso inertial parameters than without. There are some ripples in the mass estimate that occur when the robot's feet touchdown, but the estimator mostly captures the added load. For the elbow jointspace control task, we observe that an adaptive estimate is vital for the robot to maintain its ``home pose'' configuration in the presence of a time-varying load. The root-mean-square error over the two tasks, without and with adaptation, can be seen in Table \ref{tab:rmse}.

\subsection{Hardware}

The experimental procedure for hardware testing is complicated by the safety concerns that arise when fixing external loads to any powerful robot during operation. On the robot, the estimator is disabled until the robot is hoisted down to the ground and boots up into the walking state. We observed an additional bias in the residuals due to imperfections in the nominal inertial parameters, which do not capture the added inertial effects of cables, wires, and even the safety hoist on the robot. Unfortunately, this does highlight a weakness in our approach, in that we assume these nominal parameters are well-known. We are actively working on improved offline system identification routines to better characterize the nominal parameters. We have developed a calibration routine used on robot startup that computes the average residual over a recent time window, and then injects this constant bias as another term in (\ref{eq:delta-tau}). It is difficult to otherwise actively estimate this bias, as one can see by augmenting the filter state with a measurement bias $\mathbf{m}_b$ such that $\mathbf{x}' = [\boldsymbol{\pi}^\text{est}; \mathbf{m}_b]$, resulting in a process model $\mathbf{F}' = [\mathbf{I}, \mathbf{0}; \mathbf{0}, \mathbf{I}]$ and measurement model $\mathbf{H}' = [\mathbf{Y}, \mathbf{I}]$ and noting that the resultant observability matrix is rank-deficient.

\begin{figure}
\vspace{0.5\baselineskip}
     \begin{subfigure}[b]{0.32\linewidth}
         \centering
         \includegraphics[width=\linewidth]{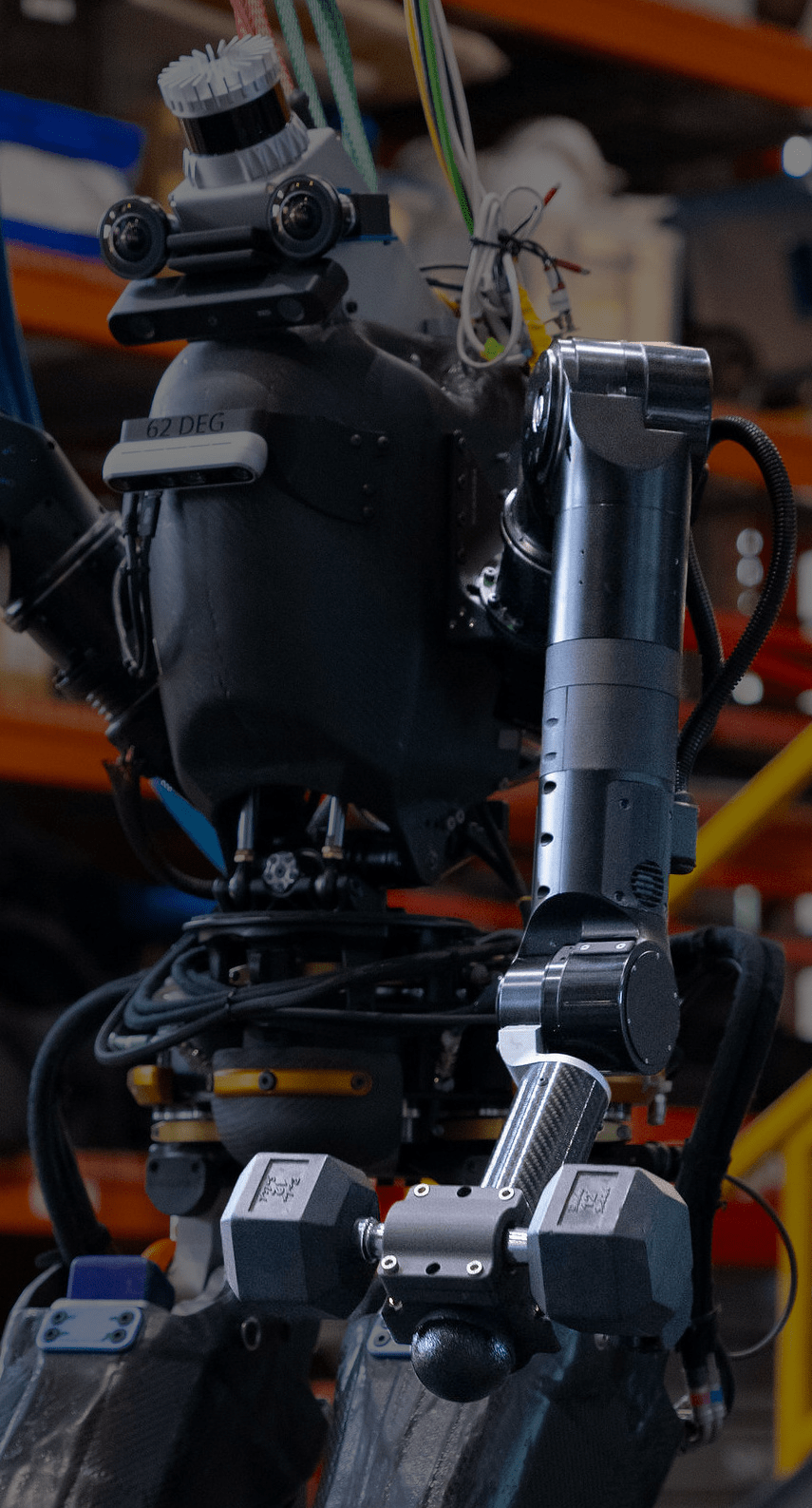}
     \end{subfigure}
     \hfill
     \begin{subfigure}[b]{0.32\linewidth}
         \centering
         \includegraphics[width=\linewidth]{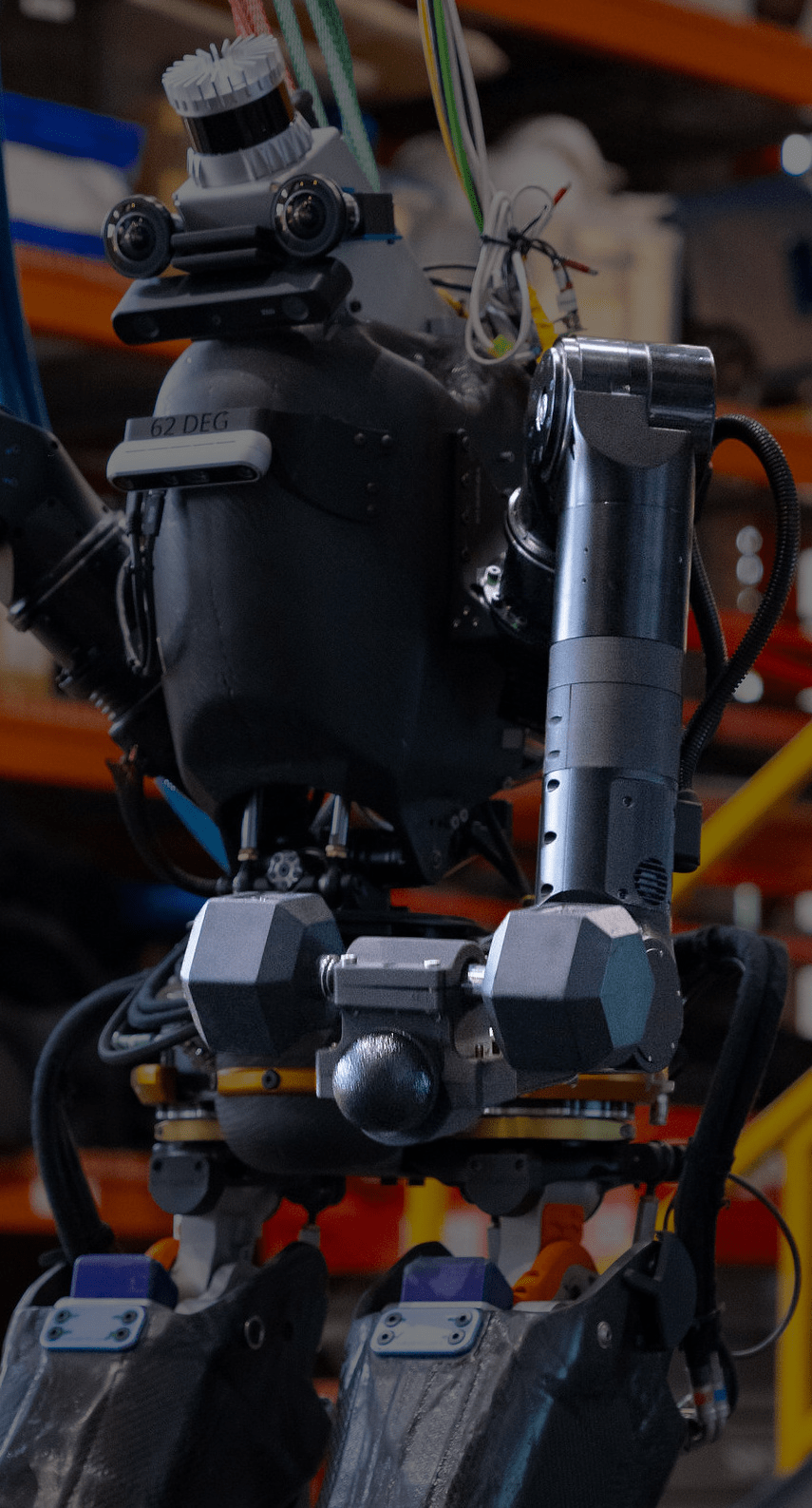}
     \end{subfigure}
     \hfill
     \begin{subfigure}[b]{0.32\linewidth}
         \centering
         \includegraphics[width=\linewidth]{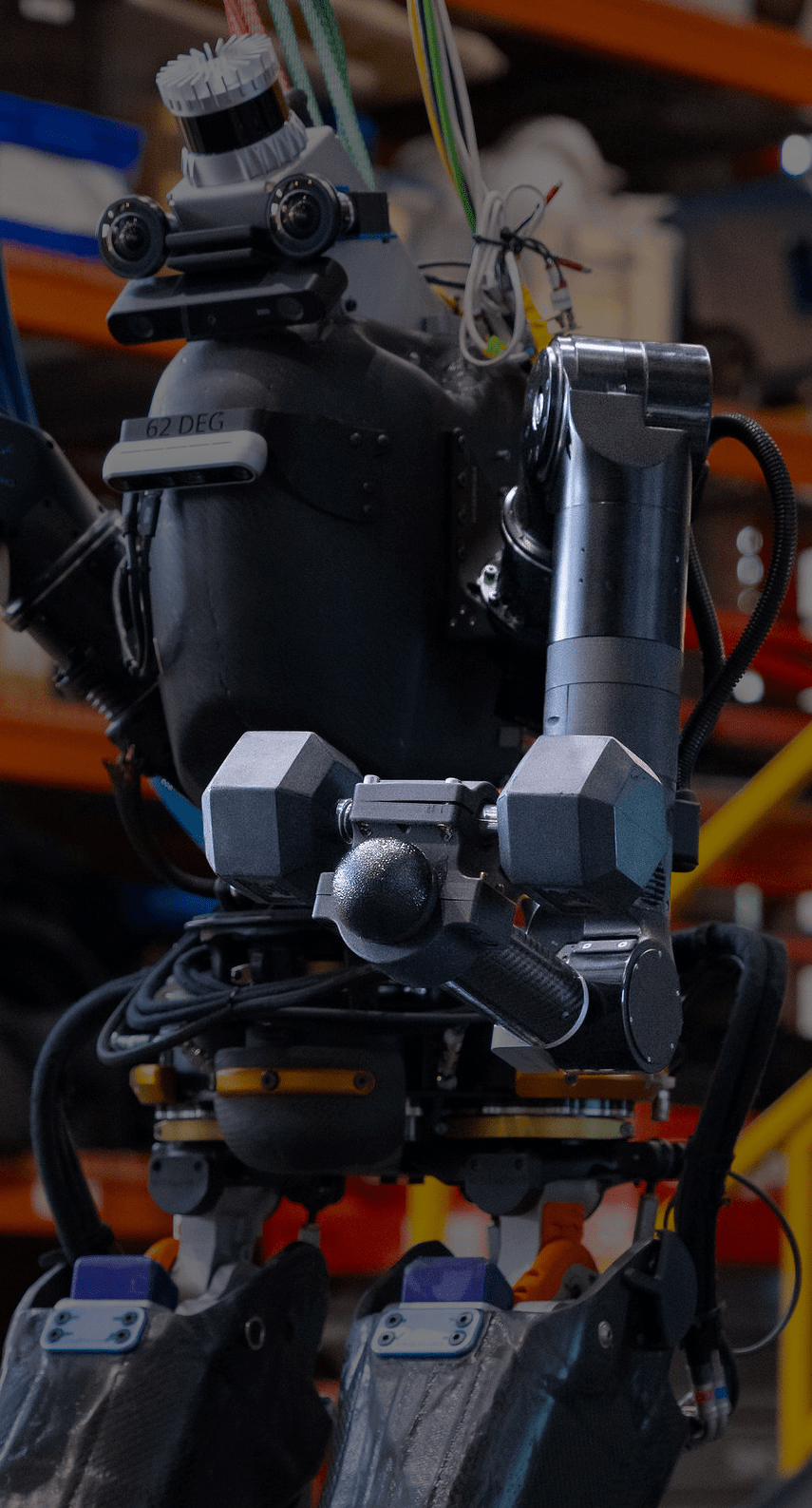}
     \end{subfigure}
        \caption{Nadia curling a 12lb dumbbell after estimating the inertial parameters of the combined dumbbell and forearm link.}
        \label{fig:curl-sequence}
\end{figure}

\begin{figure}
    \centering
    \includegraphics[width=\linewidth,height=!]{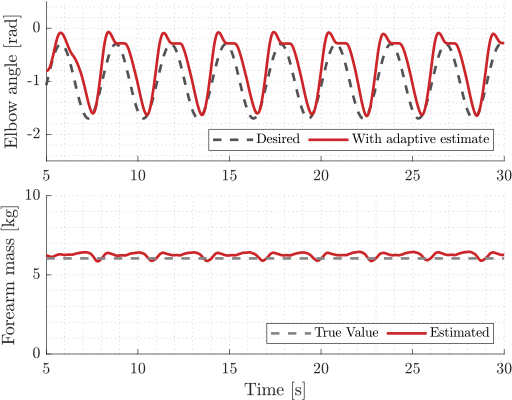}
    \caption{Elbow joint tracking and adaptive estimate of the forearm link mass during bicep curls.}
    \label{fig:hardware-curls}
\end{figure}

To test our combined estimation and control framework on hardware, we attached a 12lb dumbbell to the left arm, and then tasked the robot with performing bicep curls, as seen in Fig. \ref{fig:curl-sequence}. No prior information about the dumbbell was passed to the robot. For safety purposes, the dumbbell was attached to the robot before activation. When the estimator is first activated, the parameter estimates are fed through to the controller in a rate-limited manner. Taking mass as an example, we cap the rate at 3kg/s to prevent large initial step changes in controller objectives. The bicep curl is executed by commanding a desired elbow angle from a sine wave of 0.7rad amplitude and 0.333Hz frequency. The tracking performance of the controller and the corresponding mass estimate are shown in Fig. \ref{fig:hardware-curls}. The combined mass of the forearm link and the dumbbell is slightly overestimated, and ripples at the top and bottom of the curl. We believe this is primarily due to imperfections in other inertial parameters of the robot that are assumed known. As previously mentioned, we approximate these imperfections with a constant bias, whereas in actuality their effects are state-dependent. The elbow angle in Fig. \ref{fig:hardware-curls} generally follows the reference for the curl, but with some phase lag and a deadband-like response at the bottom of the curl. It is worth noting that without the estimate being fed into the controller, the robot cannot even achieve the home pose, limply hanging the arm down by the robot's side.  

 \begin{figure*}[t!]
 \vspace{\baselineskip}
    \centering
    \includegraphics[width=\linewidth,height=!]{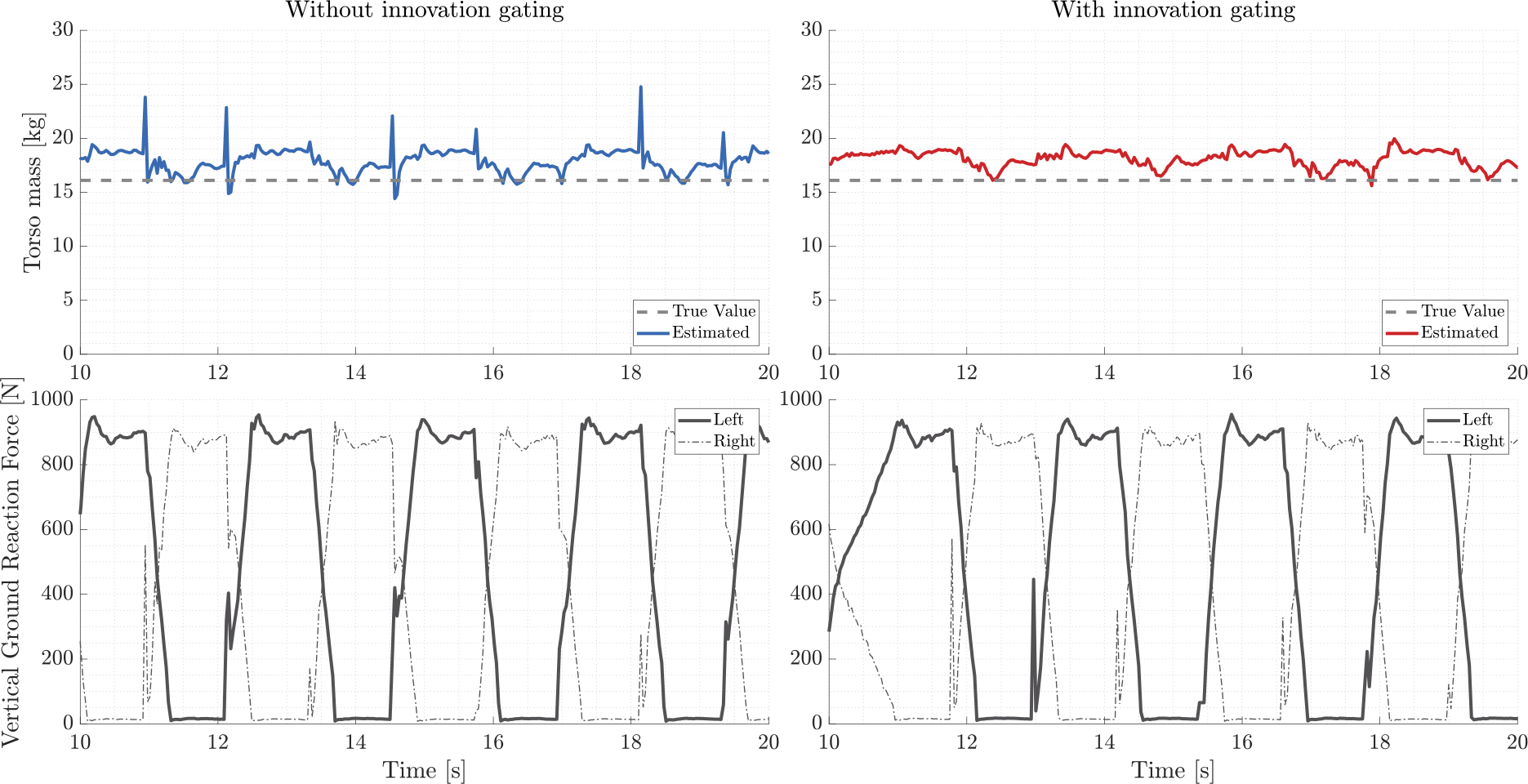}
    \caption{Comparison of torso link mass estimates before and after introducing innovation gating during hardware tests, along with the corresponding vertical ground reaction forces. We believe that the ``true value" of the torso mass drawn directly from CAD likely underestimates the mass of the link when cabling and other features are added.}
    \label{fig:hardware-gating}
\end{figure*}

When testing an early version of the estimator we noticed that significant spikes in the parameter estimates occur around contact times. During locomotion, foot impacts with the ground result in shocks in the acceleration, torque, and wrench signals, which immediately manifest in the filter residual, causing an impulse response in the filter estimate. Rather than trying to pre-process the input signals to the filter, we found the best solution was to use \textsl{innovation gating} (see \cite[Chapter 7]{grewal2014kalman} and the monograph \cite[Chapter 5.4]{bar2004estimation}). We compute the \textsl{normalized innovation} from each measurement update and reject those above a normalization threshold $\gamma$. Therefore, for a measurement to be utilized, we require:
\begin{equation}
    \Delta \tilde{\boldsymbol{\tau}}_k^\top \mathbf{S}_k^{-1} \Delta \tilde{\boldsymbol{\tau}}_k \le \gamma.
\end{equation}
There is a statistical interpretation of the normalized innovation varying as a $\chi^2$ distribution with $\gamma$ acting as the bound for a confidence interval. In our situation, we simply resorted to tuning the value online, and leave more rigorous investigation of this value, and adaptive schemes for changing it in the presence of extremely large estimate variations, to future work. The effect of this innovation gating can be seen in Fig. \ref{fig:hardware-gating}, where the robot is commanded to walk in place and the effect of the contact impulses on the torso link mass estimate is recorded. If the measurements are not gated, undesirable spikes are consistently seen in the estimate, whereas with innovation gating, the effect of the spikes is largely attenuated.

Unfortunately, we do not yet have the experimental apparatus in place to measure changes in the torso inertial parameters in a verifiable manner, as we do with the arms. However, we show that we can reliably recover a mass estimate close to the nominal mass of the robot in our URDF. We stress that this URDF value is likely to be underestimating the true mass of the link, as the URDF value is pulled directly from CAD, where the mass contributions from wires and other fixings to the torso are not captured. Therefore, we consider our estimate to be reasonable.
\section{Conclusions and Future Work}

Online adaptation to external loads such that locomotion and manipulation tasks can be satisfactorily completed will be a cornerstone requirement for humanoid robots in the field.  In this work we have introduced an estimation and control framework that facilitates online adaptation of the inertial parameters of a humanoid robot. We leveraged a recently proposed inertial parameterization that is fully physically consistent by construction--and therefore does not require any constraints--but does introduce nonlinearity, mandating that an extended Kalman Filter is used. By assuming the completion of an offline system identification phase, and that the robot will be engaging in tasks involving certain links, we showed that the problem can be reduced in size significantly, mitigating identifiability issues and permitting more efficient computation in the rigid body dynamics algorithms used inside the filter.

We demonstrated the effectiveness of the proposed filter both in simulation and on hardware, showing that Nadia is able to identify and accommodate a 12lb dumbbell being attached to the wrist link while performing bicep curls. We also showed the robot's ability to closely estimate the mass of the torso link while stepping in place and introduced an innovation gating mechanism in the EKF measurement update to reject spikes in the estimates due to contact impulses.

Several avenues exist for future work. One assumption we are keen to remove is the prior knowledge of which links are to vary inertially during operation. A sensing modality that humans possess that is frustratingly absent for this application is the sense of touch, which would allow us to localize an external load on the robot and thus inform what rigid bodies need parameter updates. Finally, we are investigating composite adaptive control approaches, where in addition to a parameter update term driven by model identification error (as done in this work), there is also a term driven by task error. While a key benefit of the estimator in this work is its independence from a specific reference model or controller, it is true that oftentimes we care more about task performance than exact parameter identification; including a parameter update term driven by task error would contribute to this objective.
\section{Acknowledgments}

We would like to thank Khizar Mohammed and Nick Kitchel for robot maintenance and experimental assistance; Bhavyansh Mishra, Geoffrey Clark, and Carlos Mastalli for insightful discussions and feedback; Achintya Mohan for graphics assistance; and the team at Boardwalk Robotics for fabricating the dumbbell mounts.

\bibliographystyle{IEEEtran}
\bibliography{IEEEabrv, references}

\end{document}